\documentclass{article}
\usepackage{ijcai25}

\usepackage{times}
\usepackage{soul}
\usepackage{url}
\usepackage[hidelinks]{hyperref}
\usepackage[utf8]{inputenc}
\usepackage[small]{caption}
\usepackage{graphicx}
\usepackage{amsmath}
\usepackage{amsfonts}
\usepackage{amsthm}
\usepackage{mathtools}
\usepackage{booktabs}
\usepackage{array}
\usepackage{algorithm}
\usepackage{algorithmic}
\usepackage[switch]{lineno}
\usepackage{multirow}
\usepackage{makecell}
\usepackage{tabu}
\usepackage[table]{xcolor}

\title{Differentiable Prompt Learning for Vision Language Models}

\author{
Zhenhan Huang$^1$
\and
Tejaswini Pedapati$^2$\and
Pin-Yu Chen$^{2}$\And
Jianxi Gao$^1$\\
\affiliations
$^1$Rensselaer Polytechnic Institute\\
$^2$IBM Research\\
}

\graphicspath{ {./figs/} } 

\definecolor{myblue}{RGB}{49, 146, 188}
\definecolor{myred}{RGB}{185, 84, 65}
\definecolor{mylightred}{RGB}{245, 90, 74}
\newcommand{\abs}[1]{\lvert#1\rvert}

\newtheorem{definition}{Definition}[section]

\usepackage{xspace}
\makeatletter
\DeclareRobustCommand\onedot{\futurelet\@let@token\@onedot}
\def\@onedot{\ifx\@let@token.\else.\null\fi\xspace}

\def\eg{\emph{e.g}\onedot} 
\def\ie{\emph{i.e}\onedot}

\makeatother

\newcommand{\answerTODO}[1][]{\textcolor{red}{\bf [TODO]}}
\newcommand{\justificationTODO}[1][]{\textcolor{red}{\bf [TODO]}}

\begin{document}

\maketitle

\begin{abstract}
Prompt learning is an effective way to exploit the potential of large-scale pre-trained foundational models. Continuous prompts parameterize context tokens in prompts by turning them into differentiable vectors. Deep continuous prompts insert prompts not only in the input but also in the intermediate hidden representations. 
Manually designed deep continuous prompts exhibit a remarkable improvement compared to the zero-shot pre-trained model on downstream tasks.
How to automate the continuous prompt design is an underexplored area, and a fundamental question arises, \textit{is manually designed deep prompt strategy optimal?}
To answer this question, we propose a method dubbed differentiable prompt learning (DPL).
The DPL method is formulated as an optimization problem to automatically determine the optimal context length of the prompt to be added to each layer, where the objective is to maximize the performance.
We test the DPL method on the pre-trained CLIP. We empirically find that by using only limited data, our DPL method can find deep continuous prompt configuration with high confidence.
The performance on the downstream tasks exhibits the superiority of the automatic design: our method boosts the average test accuracy by \textbf{2.60 \%} on 11 datasets compared to baseline methods. 
Besides, our method focuses only on the prompt configuration (\ie context length for each layer), which means that our method is compatible with the baseline methods that have sophisticated designs to boost the performance. 
The DPL method can be deployed to large language models or computer vision models at no cost.
\end{abstract}


\section{Introduction}
Parameter-Efficient fine-tuning (PEFT) offers a paradigm to adapt pretrained model to downstream tasks using a small set of trainable parameters. Among the PEFT methods, the prompt learning can perform few-shot learning or even zero-shot learning to adapt the pre-trained models to new scenarios. Using appropriate prompts, both text prompts \cite{brown2020language,lester2021power,schick2020s,shin2020autoprompt,gao2020making} and visual prompts \cite{bar2022visual,shtedritski2023does,chen2023understanding,tsai2024convolutional}, to query foundational models have shown great potential. At the same time, the selection of context tokens in prompts has a pronounced effect on the performance of the prompt learning methods \cite{zhou2022learning,schick2021few,schick2020exploiting,haviv2021bertese}. 
Continuous prompts use differentiable vectors that are in the embedding space of the foundation model.
Continuous prompts use trainable parameters that are learned during the fine-tuning process.
It avoids the time-consuming trial-and-error process using discrete context tokens.
Besides, the continuous prompts are not constrained to be the embeddings of word tokens in the vocabulary.
The selection for the embeddings of context tokens becomes continuous.

Deep continuous prompts add prompts to multiple layers. This method has shown superior performance compared to only adding continuous prompts to the input in both vision \cite{jia2022visual,zhu2023visual,zhu2023visual,yoo2023improving} and language fields \cite{lester2021power,li2021prefix,liu2021p,liu2023gpt}. There are two hyperparameters: the context length of continuous prompts $c_p$ and prompt depth $\ell_p$. Let the hidden dimension be $d$, continuous prompts $\mathbf{E} \in \mathbb{R}^{c_p \times d}$ are inserted to the input to the neural network layers up to $\ell_p$ layers. Since $\mathbf{E}$ are automatically determined during the fine-tuning process, a natural question is: \textit{can we automatically determine $c_p$ and $\ell_p$?} 
Recently, it has been found that training merely part of deep neural network layers can achieve a performance comparable to or even better than training all layers. This line of work indicates that there is a subset of layers in the pre-trained model, depending on the distribution shift between the pre-training dataset and fine-tuning dataset, whose parameters might be close to a minima for downstream tasks 
\cite{lee2022surgical,lodha2023surgical,panigrahi2023task,vettoruzzo2024advances}.


In the prompt learning, we postulate that there might be some layers that do not need deep continuous prompts or only need continuous prompts with shorter context length compared to rest of layers. In the existing prompt learning works, however, a fixed number $c_p$ is used for each layer. To examine our postulation, we design our method to insert continuous prompts with different context lengths. 
The context lengths of continuous prompts are determined using a differentiable formulation. If a layer has the best context length found to be 0, we do not add continuous prompts to this layer.
We name the proposed method \emph{differentiable prompt learning} (DPL). The few-shot learning experiments show that the DPL method can find continuous prompt configurations, i.e., the context length and depth of continuous prompts inserted to the input of each layer. The performance of downstream fine-tuning over 11 datasets shows the superiority of the proposed method.

In summary, our DPL method has the following contributions:
\begin{itemize}
    \item The DPL method automates the continuous prompt learning process. It automatically determines the continuous prompts and associated hyperparameters including the context length of the continuous prompt inserted for each neural network layer and prompt depth.
    \item The DPL method removes the constraint in the manually designed continuous prompt methods that the context length for each layer is fixed. The heterogeneous design of inserting continuous prompts enables a more flexible design for the prompt learning. The method is simple and focuses only on continuous prompts, which means it can be combined with sophisticated designs in existing prompt learning methods.
    \item We find the optimal deep continuous prompt configuration is dataset-dependent. By tailoring deep continuous prompts for each dataset, the DPL method can achieve the performance better than manually designed prompt learning methods. 
\end{itemize}

\section{Background and Related Work}

\paragraph{Prompt Learning} Prompt learning offers an efficient way to adapt pre-trained foundational models to downstream tasks. This technique is of great interest in both vision \cite{wang2022learning,wang2022dualprompt} and natural language \cite{jiang2020can,shin2020autoprompt}. CoOp \cite{zhou2022learning} inserts continuous prompts to the input of the vision-language model. The visual prompt tuning \cite{jia2022visual} proposes the deep continuous prompts to vision transformer \cite{dosovitskiy2020image}. MaPLe \cite{khattak2023maple} combines the ideas from these two works by using deep continuous prompts in both the text branch and the image branch. Similar to CoOp, CoCoOp, PLOT and ProGrad insert continuous prompts only in the input. CoCoOp \cite{zhou2022conditional} inserts the continuous prompts conditioning on the input images. PLOT \cite{chen2022plot} uses the computationally costly iterative algorithm to compute the transport plan for aligning word tokens and image patches. ProGrad \cite{zhu2023prompt} uses the gradient-aligned knowledge distillation to avoid the overfitting problem. All of the above-mentioned methods use fixed context length as opposed to our proposed DPL method which adapts the context length for each layer dynamically based on the dataset.

\paragraph{Neural Architecture Search} Neural architecture search (NAS) automatically determines the architecture by minimizing the objective function. One-shot NAS is an efficient way to search neural architectures. In the one-shot NAS, a supernet \cite{saxena2016convolutional,bender2018understanding,pham2018efficient,liu2018darts} is introduced comprising all possible architecture in the search space. After training the supernet, a subnet is uniquely determined as the neural architecture determined by NAS.

\subsection{Revisiting Differentiable NAS}
Differentiable NAS \cite{liu2018darts,xu2019pc,dong2019searching,liang2019darts+,zela2019understanding,chu2020fair,yan2021zeronas} uses a cell-based search space. A cell is represented by a directed acyclic graph (DAG) $\mathcal{G}(\mathcal{V}, \mathcal{E})$. Each node is a hidden representation and each directed edge is an operation transforming the hidden representation. The supernet incorporates all the operations. Let $\mathcal{O}$ be a set of candidate operations. The categorical choice of a particular operation is relaxed by the softmax over all possible operations:
\begin{equation}
    \bar{o}^{(i, j)}(\mathbf{x}) = \sum_{o \in \mathcal{O}} \frac{\exp(\alpha_{o}^{(i, j)})}{\sum_{o^{\prime}} \exp(\alpha^{(i, j)}_{{o}^{\prime}})} o(\mathbf{x}) \;,
\end{equation}
where $\alpha^{(i, j)}$ are trainable parameters that determine the searched operation between the latent representation $i$ and $j$. The goal of the search process is to replace $\bar{o}^{(i, j)}$ with the most likely operation. The optimal neural architecture is obtained after the search process.

Inspired by differentiable NAS methods, we relax the categorical selection on the context lengths of continuous prompts to make the search space continuous. We use the search space to determine the optimal prompt configuration. The continuous prompts are trained from scratch. Existing vision-language prompt learning methods use the deep prompt method \cite{jia2022visual} where continuous prompts are added to transformer blocks and removed after the self-attention \cite{vaswani2017attention}:
\begin{equation}
    [\mathbf{x}^{(l)}, \mathbf{E}^{(l)}] = f^{(l)}([\mathbf{x}^{(l-1)}, \mathbf{E}^{(l-1)}]) \;.
\end{equation}
Here we denote $l$-th transformer block as $f^{(l)}$. Different options for adding continuous prompts cannot be mixed as the dimension of the input $[\mathbf{x}^{(l-1)}, \mathbf{E}^{(l-1)}]$ is different due to different context lengths. We solve this problem by using cross-attention. Details are reported in the following section.

\section{Differentiable Prompt Learning}
\label{sec:dpl}

We use the pre-trained CLIP model \cite{radford2021learning} in our experiments. The CLIP model is pre-trained on over 400 million image-text pairs. 
The DPL method has two stages: the searching stage and the training stage. The goal of the searching stage is to determine the context length of the continuous prompt to be added to each transformer block of the CLIP model. The best prompt configuration is used in the training stage. Figure \ref{fig:illust} shows the proposed method.

\subsection{Searching Stage}

We prepend continuous prompts $\{ \mathbf{E}^{(l)} \}$ to inputs to transformer blocks $\{ \mathbf{x}^{(l)} \}$ in the text branch and image branch, where $1 \le l \le \ell, l \in \mathbb{N}^{+}$. The concatenated inputs are fed into transformer blocks $f(\cdot)$. \textit{w.l.o.g}, for the $l$-th layer, there are $t$ options for adding continuous prompts of different context length $\mathbf{E}^{(l)}_i \in \mathbb{R}^{c_{i} \times d}$, where $1 \le i \le t, i \in \mathbb{N}^{+}$, to the input $\mathbf{x}^{(l)} \in \mathbb{R}^{c_l \times d}$. $c_{i}$ is the context length for $i$-th option, $d$ is the hidden dimension.
For the text branch, the hidden dimension is $d = d_{\rm txt}$. For the image branch, the hidden dimension is $d = d_{\rm img}$. Continuous prompts in the text branch are independent of those in the image branch. The output $\mathbf{h}^{(l)}$ of the transformer block is:
\begin{equation}
    \mathbf{h}^{(l)}_i = \begin{cases}
        f^{(l)}([\mathbf{x}^{(l-1)}]) \;, & i = 1 \\
        f^{(l)}([\mathbf{E}_{i}^{(l)}, \mathbf{x}^{(l-1)}]) \;. & 1 < i \le t
    \end{cases}
\end{equation}
$i = 1$ accounts for the case where no continuous prompt is added. Different from the traditional transformer \cite{vaswani2017attention} where self-attention mechanism is applied, we use the cross-attention mechanism in transformer blocks:
\begin{equation}\label{eq:cross_att_qkv}
    \mathbf{Q} = \mathbf{w}_q^T\mathbf{x} ,\  \mathbf{K} = \mathbf{w}_k^T[\mathbf{E}_{i}^{(l)}, \mathbf{x}^{(l-1)}] , \ \mathbf{V} = \mathbf{w}_v^T[\mathbf{E}_{i}^{(l)}, \mathbf{x}^{(l-1)}] \;.
\end{equation}
\begin{equation}\label{eq:cross_att_multi}
    \begin{split}
    & \text{Cross Attention} = \text{Softmax}(\frac{\mathbf{Q}\mathbf{K}^T}{\sqrt{d}})\mathbf{V} , \\
    & \text{ where } \mathbf{Q} \in \mathbb{R}^{c_l \times d}, \  \mathbf{K} \in \mathbb{R}^{(c_i+c_l)\times d}, \ \mathbf{V} \in \mathbb{R}^{(c_i+c_l)\times d} \;.
    \end{split}
\end{equation}

\begin{figure}[t]
    \centering
    \includegraphics[width=\linewidth]{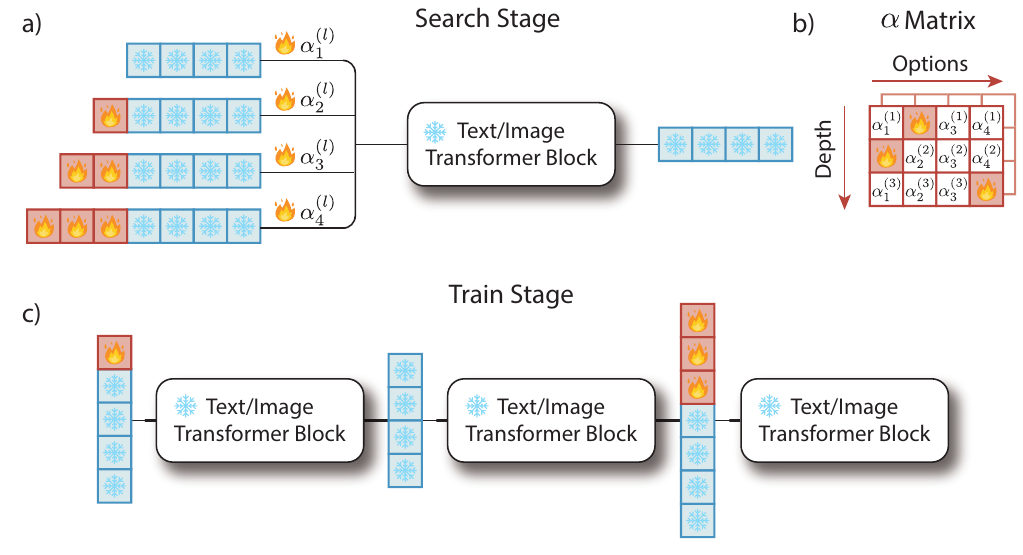}
    \caption{(a) In the searching stage, continuous prompts with different context lengths (shown in \textcolor{myred}{red} color) are added to the original context tokens (shown in \textcolor{myblue}{blue} color) as the input to transformer blocks in the text branch or image branch. The outputs of transformer blocks are used as the original context tokens for the next transformer blocks. $\{\alpha^{(l)}_i\}$ are differentiable parameters to control the contribution of different prompt options. (b) After the searching stage, two $\alpha$ matrices are obtained to indicate the selection of the search algorithm for differentiable context tokens in the language branch and the image branch. (c) In the training stage, prompt learning is conducted using the differentiable context token setting determined by the search algorithm.}
    \label{fig:illust}
\end{figure}

We use parameters $\{\beta^{(l)}_i\}_{i=1}^{t}$ to control the contribution of different options to the output. In other words, the mixing weight for the hidden representation $\mathbf{h}^{(l)}$ is parameterized by $\beta^{(l)}_i$:
\begin{equation}\label{eq:mix}
    \mathbf{x}^{(l)} = \sum_{i = 1}^t \beta^{(l)}_i \mathbf{h}^{(l)}_i \;.
\end{equation}
The Equation \ref{eq:mix} indicates that the output of each transformer block is a combination of all possible context lengths, \ie $[0, t]$. $\{\beta_i^{(l)}\}_{i=1}^{t}$ is computed by differentiable parameters $\{\alpha^{(l)}_i\}_{i=1}^{t}$:
\begin{equation}
    \beta_i^{(l)} = \frac{\exp(\alpha^{(l)}_i)}{\sum_{m=1}^t \exp(\alpha^{(l)}_{m})} \;.
\end{equation}

Figure \ref{fig:illust} (a) shows the illustration of the searching stage. Same as conventional prompt learning methods, the parameters of the transformer blocks of the pre-trained model are frozen. During the searching process, updated parameters are continuous prompts and $\alpha$ parameters. The goal of updating continuous prompts is to minimize the training loss $\mathcal{L}_{\rm train}$ while that of updating $\alpha$ parameters is to minimize the validation loss $\mathcal{L}_{\rm val}$. This implies that the searching stage is a bilevel optimization problem \cite{anandalingam1992hierarchical,colson2007overview} with the upper level parameter $\alpha$ and the lower level parameter $\mathbf{E}$:
\begin{align}
    \underset{\alpha}{\text{min}} \quad & \mathcal{L}_{\rm val}(\mathbf{E}^*(\alpha), \alpha) \\
    \text{s.t.} \quad & \mathbf{E}^*(\alpha) = \underset{\mathbf{E}}{\text{argmin}} \ \mathcal{L}_{\rm train}(\mathbf{E}, \alpha) .
\end{align}

Algorithm \ref{alg:search_dpl} shows the searching stage. The input is the pre-trained vision-language model and two $\alpha$ matrices (we call them search spaces). Two $\alpha$ matrices are randomly initialized before training. After convergence, the $\alpha$ matrix $\mathbf{A}^{\alpha} \in \mathbb{R}^{\ell \times t}$ is obtained as shown in Figure \ref{fig:illust} (b). The row dimension is associated with the depth and the column dimension is related to $t$ options. The column index of the highest $A^{\alpha}_{im}$ ($1 \le m \le t, m \in \mathbb{N}^{+}$) for the $i$-th transformer block determines the context length of the continuous prompts added to that block.

\begin{algorithm}[htb]
\caption{Searching stage for vision-language models}\label{alg:search_dpl}
\begin{algorithmic}[1]
  \STATE {\bfseries Input}: A pre-trained model and two $\alpha$ matrices $\mathbf{A}^{\alpha} \in \mathbb{R}^{\ell \times t}$ with randomly initialized weights.
  \WHILE{not converged}   
    \STATE Update $\mathbf{A}^{\alpha}$ by descending $\nabla_{\mathbf{A}^{\alpha}} \mathcal{L}_{\rm val}(\mathbf{E}, \mathbf{A}^{\alpha})$.
    \STATE Update continuous prompts in both text branch and image branch by descending $\nabla_{\mathbf{E}} \mathcal{L}_{\rm train}(\mathbf{E}, \mathbf{A}^{\alpha})$.
  \ENDWHILE
  \FOR {$i = 1$ to $\ell$}
    \STATE $A^{\alpha}_{ik} = \max_{m}A^{\alpha}_{im}$, $k$ determines the context length of continuous prompts for the $i$-th block in the best prompt configuration.
  \ENDFOR
  \STATE {\bfseries Output}: Prompt configuration for the image branch and the text branch.
\end{algorithmic}
\end{algorithm}

We use the term \emph{supprompt} to represent the combination of the continuous prompts added to the pre-trained model and an $\alpha$ matrix in the searching stage. For vision-language models, there are two sets of supprompts for the image and text branches. After the search, the best context length is identified for each layer and the resulting model is then trained from scratch. We use the term \emph{subprompt} to denote the final best context length for each layer that is used in the final model. Terminologies are summarized in Appendix \ref{append:terminology}.

Similar to the differentiable NAS where the supernet incorporates all candidate operations, our method creates a supprompt that contains all prompt configurations in the search space. Incorporation of all possibilities inevitably leads to a relatively large supprompt compared to the subprompt, which can increase the computational cost in the searching stage. After the optimal prompt configuration is determined, the computational complexity is small compared to the deep continuous prompt method such as MaPLe. The complete comparison is reported in the Appendix \ref{append:complexity}.

\subsection{Training Stage}

We denote $\mathbf{A}^{\alpha}_{\rm txt}$ for the $\alpha$ matrix of the text branch and $\mathbf{A}^{\alpha}_{\rm img}$ for that of the image branch. The column index of the maximum value in each row of $\alpha$ matrix determines the context length of the inserted continuous prompts, i.e. $A^{\alpha}_{ik} = \max_{m} A^{\alpha}_{im}$. After the context length of continuous prompts is determined, we fine-tune the model on various downstream datasets. Figure \ref{fig:illust} (c) shows the training stage. The fine-tuning process is the same as the conventional prompt learning method: model parameters are frozen and only continuous prompts are differentiable.

In the image branch, the input $\mathbf{x} \in \mathbb{R}^{C \times H \times W}$ is patchified and projected to produce patch token embeddings \cite{dosovitskiy2020image}. A learnable classification token embedding $\mathbf{u}_{\rm cls} \in \mathbb{R}^{1 \times d_{\rm txt}}$ is added to the patch token embedding $\tilde{\mathbf{x}}_{\rm img} = [\mathbf{u}_{\rm cls}, \mathbf{u}_{1}, \ldots, \mathbf{u}_{n_{\rm img}}]$. In the text branch, a text template such as \texttt{a photo of a [class]} is tokenized and embedded to generate text token embeddings that are formulated as $\tilde{\mathbf{x}}_{\rm txt} = [\mathbf{v}_{\rm SOS}, \mathbf{v}_{1}, \ldots, \mathbf{v}_{n_{\rm txt}}, \mathbf{v}_{k}, \mathbf{v}_{\rm EOS}]$, where $\mathbf{v}_{\rm SOS}$ is the start token embedding, $\mathbf{v}_{\rm EOS}$ is the end token embedding, and $\mathbf{v}_k$ is the embedding corresponding for the class name. $\tilde{\mathbf{x}}_{\rm txt}$ and $\tilde{\mathbf{x}}_{\rm img}$ are combined with continuous prompts and fed to transformer blocks in the text branch and image branch for obtaining text embedding $\tilde{\mathbf{g}}$ and image embedding $\tilde{\mathbf{f}}$, respective. Within each transformer block, we use the cross attention mechanism as shown in Equation \ref{eq:cross_att_qkv} and \ref{eq:cross_att_multi}. The output logits are based on the similarity between $\tilde{\mathbf{g}}$ and $\tilde{\mathbf{f}}$:
\begin{equation}
    \hat{\mathbf{y}} = \frac{\exp(\text{sim}(\tilde{\mathbf{f}}, \tilde{\mathbf{g}}) / \tau)}{\sum_{c \in \mathcal{C}} \exp(\text{sim}(\tilde{\mathbf{f}}, \tilde{\mathbf{g}}_c) / \tau)} \;,
\end{equation}
where $\tau$ is the temperature parameter. The loss $\mathcal{L}_{\rm de}$ computes the deviation of the prediction from the ground truth labels:
\begin{equation}
    \mathcal{L}_{\rm de} = \mathbb{E}_{(\mathbf{x}, y) \sim \mathcal{D}} \ \mathcal{F}(\hat{\mathbf{y}}, y) \;,
\end{equation}
where $\mathcal{F}(\cdot)$ is the loss function. We use the cross entropy loss for calculating $\mathcal{L}_{\rm de}$. Continuous prompts are updated by descending the training loss $\nabla_{\mathbf{E}} \mathcal{L}_{\rm train}(\mathbf{E})$.

In addition to the vanilla DPL, we add knowledge distillation in the training stage. Specifically, we use Kullback-Leibler (KL) divergence \cite{hinton2015distilling} between the model prediction $\hat{\mathbf{y}} = p(\mathbf{x})$ and the zero-shot pre-trained model prediction $p_{\rm zs}(x)$:
\begin{equation}
    \mathcal{L}_{\rm kl} = -\sum_{\mathbf{x} \in \mathcal{X}} p_{\rm zs}(\mathbf{x}) \log(\frac{p(\mathbf{x})}{p_{\rm zs}(\mathbf{x})}) \;.
\end{equation}
The total loss $\mathcal{L}_{\rm total}$ is computed by:
\begin{equation}
    \mathcal{L}_{\rm total} = \mathcal{L}_{\rm de} + \lambda \mathcal{L}_{\rm kl} \;,
\end{equation}
where $\lambda$ is a hypereparameter. The gradient descent of continuous prompts uses $\mathcal{L}_{\rm total}$ to update parameters.

\section{Experiments}

\subsection{Datasets and Experiment Setup}

\paragraph{Datasets} We evaluate the DPL method on 11 datasets: Caltech101 \cite{fei2004learning} and ImageNet \cite{deng2009imagenet} for the generic object classification, DescribableTectures \cite{cimpoi2014describing} for the texture classification, EuroSAT \cite{helber2019eurosat} for the satellite image classification, FGVCAircraft \cite{maji2013fine}, Food101 \cite{bossard2014food}, OxfordFlowers \cite{nilsback2008automated}, OxfordPets \cite{parkhi2012cats}, and StanfordCars \cite{krause20133d} for the fine-grained image recognition, UCF101 \cite{soomro2012ucf101} for the action classification, and SUN397 \cite{xiao2010sun} for the scene recognition.

\paragraph{Baselines} We compare the proposed method with CoCoOp \cite{zhou2022conditional}, PLOT \cite{chen2022plot}, MaPLe \cite{khattak2023maple} and ProGrad \cite{zhu2023prompt}. The original implementation of PLOT uses ResNet \cite{he2016deep} as the backbone model. For a fair comparison, we replace the backbone model of ResNet with the transformer model. In addition to the prompt learning methods, we examine the performance of the zero-shot CLIP (ZS CLIP) and that of training a linear classifier given image and text representations by CLIP (Linear probe) \cite{radford2021learning}.

\paragraph{Experiment Details} We use the pretrained ViT-B/16 CLIP model \cite{radford2021learning} in the few-shot learning.
In both training stage and searching stage, we use the same hyperparameters except for the number of epochs. The number of epochs in the searching stage is 60 while that for the training stage is 40. The batch size is 4 and we use stochastic gradient descent (SGD) to optimize continuous prompts. In the searching stage, two $\alpha$ matrices are optimized using SGD strategy. Learning rate is $3.5 \times 10^{-3}$. Experiments are conducted using a single NVIDIA A40 GPU.

\subsection{Determining Context Lengths of Continuous Prompts}

We use the few-shot learning setting in the searching stage. The number of shots is the same for the searching and training stages. The number of shots is 16. The results of using 8/4/2/1 shots are shown in Appendix \ref{append:few_shot}. The evolution of $\alpha$ matrices is visualized to examine the convergence process.
The evolution of $\alpha$ matrices using 11 different datasets is reported in the Appendix \ref{append:alpha_evolve}. We define special $\alpha$ matrices formally:
\begin{definition}[single-dominant]
An $\alpha$ matrix $\mathbf{A}^{\alpha} \in \mathbb{R}^{\ell \times t}$ is single-dominant if $\ \forall i \in \mathbb{N}^{+}, 1 \le i \le \ell$, $\exists j \in \mathbb{N}^{+}, 1 \le j \le t$ s.t. $A^{\alpha}_{ij} \gg A^{\alpha}_{ik}$, where $k \in \mathbb{N}^{+}, 1 \le k \le t, k \ne j$.
\end{definition}

\begin{figure*}[htb]
    \centering
    \includegraphics[width=.92\linewidth]{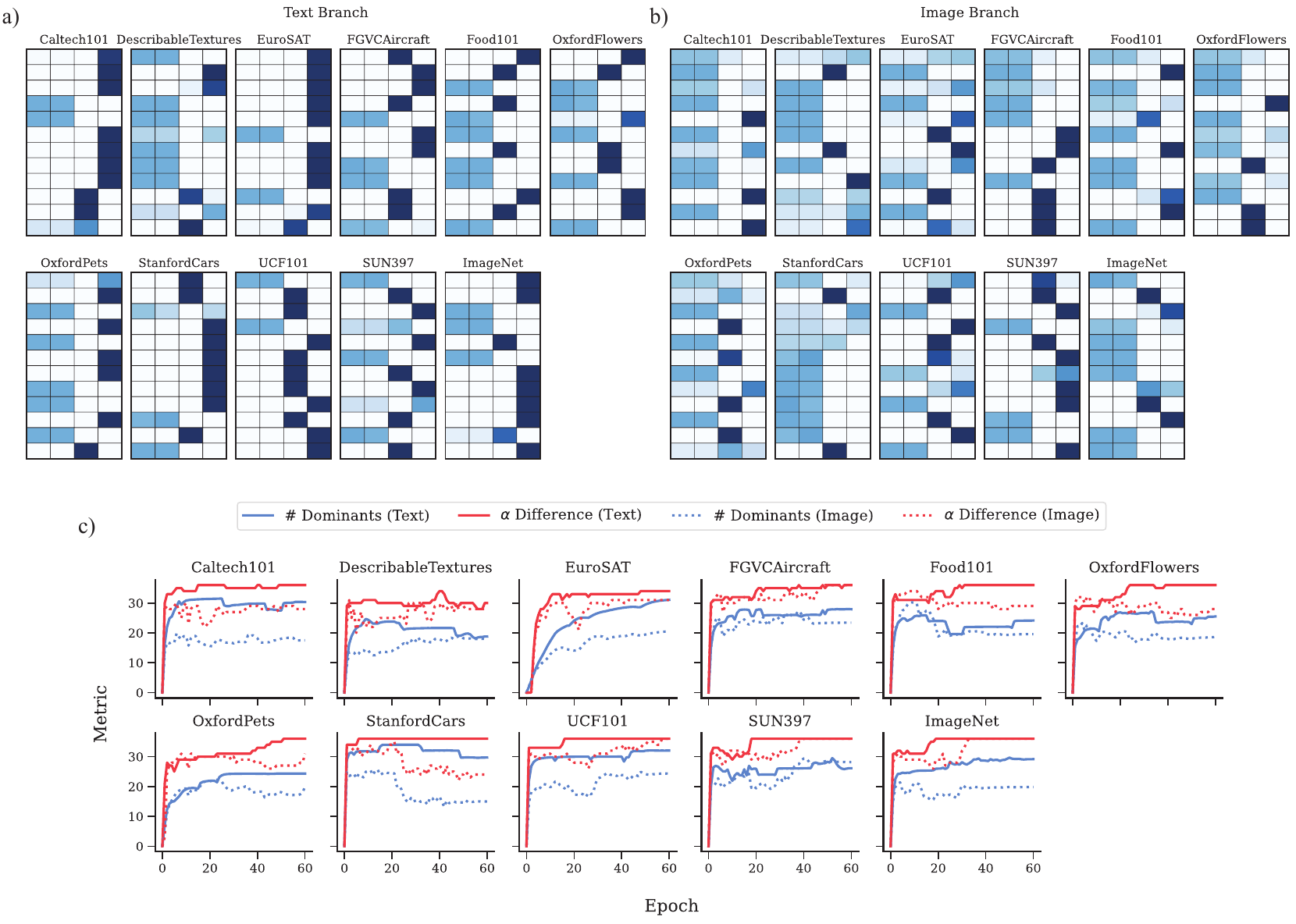}
    \caption{(a) $\alpha$ matrices for the text branch. (b) $\alpha$ matrices for the image branch. $\alpha$ matrices are obtained at the epoch of 60. The row dimension is related to the context length of added continuous prompts. The column dimension is related to model depth, i.e. the number of transformer blocks. (c) The evolution of the $\alpha$ difference and the number of dominants. As the number of training epochs increases, the $\alpha$ matrix gradually converges and the number of dominants increases to the converged value.}
    \label{fig:alpha_combine}
\end{figure*}

The single-dominant $\alpha$ matrix means that the searching algorithm has a high confidence in the searched subprompts. This is beneficial to the training stage as the determined $\alpha$ matrix is robust against the fluctuation of the matrix in the updating process.
We use the alpha value at each row with the largest value (i.e. $\text{argmax}_j \alpha_{ij}$) to determine the context length of continuous prompts added in the training stage. Hence, the training stage does not require the $\alpha$ matrix to be single-dominant. When $\alpha$ matrix is not single-dominant, the decision on adding continuous prompts is not robust. For example, in the case where two $\alpha$ values are close:
\begin{equation}
    \exists \epsilon \quad \text{s.t.} \ \epsilon > 0 ,\ \epsilon \ll \abs{A^{\alpha}_{im}} ,\ \epsilon \ll \abs{A^{\alpha}_{in}} ,\ \abs{A^{\alpha}_{im} - A^{\alpha}_{in}} < \epsilon \;,
\end{equation}
the fluctuation can change the order in the comparison of $\alpha$ values. At the epoch of 60, if we use a threshold value to filter out small $\alpha$ values, $\alpha$ matrices become essentially sparse matrices.

Figure \ref{fig:alpha_combine} (a) and (b) shows $\alpha$ matrices at the epoch of 60 for all 11 datasets. Compared to the image branch, there are considerably more single-dominant $\alpha$ matrices. Although we use 16-shots learning, the searching results indicate that the searching algorithm has pretty high confidence for the searched prompt configuration. $\alpha$ matrices at the last epoch are used in the training stage for the prompt learning. To quantify the evolution process of $\alpha$ matrices, we firstly define $\alpha$ difference:
\begin{equation}
    \begin{split}
    & \alpha \text{ Difference} \coloneqq \sum_{i}\sum_{j } \abs{\text{Softmax}(A^{\alpha}_{ij}) - \text{Softmax}(A^{\alpha}_{ik})}, \\
    & \qquad \text{where}\ A^{\alpha}_{ik} = \max_m(A^{\alpha}_{im}) \;.
    \end{split}
\end{equation}
Apparently, single-dominant $\alpha$ matrix has a high $\alpha$ difference value. Figure \ref{fig:alpha_combine} (c) shows the variation of the $\alpha$ difference. At the early training stage, the $\alpha$ difference increases drastically. there is a fluctuation when the $\alpha$ difference is close to the converged value. The fluctuation comes from two parts: the first part is related to the variation of the largest $\alpha$ value in each row. Even though the $\alpha$ matrix is single dominant, the largest $\alpha$ value is subjected to the variation in the gradient descent. Owing to the softmax operation, the contribution of the largest $\alpha$ value and that of the rest $\alpha$ values in a row is not independent, hence we call this \emph{intra factor}. The second part is the fluctuation of $\alpha$ values except the largest ones. We call it \emph{inter factor}.

\begin{table*}[t]
    \caption{Test accuracies of DPL and baseline methods on 11 datasets using few-shot learning. Results are averaged over 3 runs. The DPL method shows a pronounced advantage compared to baseline methods. Combining the DPL method with knowledge distillation can further boost the performance.}
    \label{table:subnet_perf}
    \resizebox{\linewidth}{!}{
    \begin{tabular}{ccccccccccccc}
    \hline
    \multirow{2}*{Method} & \multicolumn{12}{c}{Test Accuracy $\uparrow$} \\
    \cline{2-13}
     & Caltech101 & DTD & EuroSAT & Aircraft & Food101 & Flowers & Pets & Cars & UCF & SUN397 & ImageNet & Average\\
    \hline
    ZS CLIP & 87.20 & 42.34 & 37.57 & 17.29 & 77.30 & 66.18 & 85.79 & 55.63 & 61.45 & 58.73 & 58.77 & 58.93 \\
    Linear Probe & 90.44 & 64.02 & 82.68 & 36.36 & 70.13 & 95.01 & 76.36 & 70.13 & 73.70 & 65.17 & 54.33 & 70.76 \\
    \hline
     CoCoOp  & 95.10 & 63.63 & 74.10 & 33.67 & 87.37 & 89.97 & 93.53 & 72.30 & 76.97 & 72.67 & 71.17 & 75.50 \\
     PLOT    & 93.70 & 70.90 & 84.03 & 34.93 & 78.13 & \textbf{97.27} & 88.20 & 68.10 & 72.23 & 69.40 & 72.23 & 75.37 \\
     ProGrad & 95.63 & 66.27 & 82.03 & 41.30 & 86.70 & 95.33 & 93.10 & 81.23 & 81.60 & 75.13 & 72.27 & 79.14 \\
     MaPLe   & 95.10 & 67.27 & 86.40 & 37.07 & \textbf{87.43} & 94.27 & \textbf{93.63} & 74.87 & 80.37 & 74.73 & 72.03 & 78.47 \\
     DPL    &  \textbf{95.80} & \textbf{71.50} & 92.30 & 48.57 & 82.57 & 96.63 & 92.03 & 82.70 & \textbf{84.10} & \textbf{79.83} & 72.80 & 81.71 \\
     DPL + KD & 95.73 & 70.90 & \textbf{92.47} & \textbf{49.43} & 82.80 & 96.80 & 91.83 & \textbf{82.83} & 83.77 & 79.63 & \textbf{73.03} & \textbf{81.74} \\
     \hline
    \end{tabular}}
\end{table*}

The number of dominants describes the number of pairs between the largest $\alpha$ values and the rest of $\alpha$ values in the same row:
\begin{equation}
    \begin{split}
    & \text{\# Dominants} \coloneqq \big\lvert\{(A^{\alpha}_{ij}, A^{\alpha}_{ik}) \mid 1 \le i \le \ell, 1 \le j \le t, \\
    & \qquad j \ne k, A^{\alpha}_{ik} = \max_m(A^{\alpha}_{im}), A^{\alpha}_{ik} \gg A^{\alpha}_{ij}\} \big\rvert
    \end{split}
\end{equation}
The ideal converged $\alpha$ matrix has the number of dominants $\text{\# Dominants} = \ell(t-1)$.

\subsection{Prompt Learning Based on Alpha Matrix}

After determining the context length of added continuous prompts, we add them to the pre-trained CLIP model and conduct the prompt learning in a supervised fashion. We use the few-shot learning and the number of shots is 16. The $\alpha$ matrix in the text branch might be different from that in the image branch as shown in Appendix Figure \ref{fig:alpha_mat_all}, and there is no coupling function \cite{khattak2023maple,zang2022unified} between added continuous prompts in the two branches. Table \ref{table:subnet_perf} shows the performance comparison on downstream datasets. Note that the direct comparison of zero-sho CLIP with other methods is not fair as it does not require any training.

Our proposed method shows a pronounced advantage compared to baseline methods. It boosts the average accuracy by $2.60\%$ test accuracy. The proposed method shows unfavorable performance on the Food101 dataset. When comparing the zero-shot CLIP method with the linear probe method which adds a linear classifier on top of pre-trained CLIP model, there is a remarkable performance drop ($77.30\% \rightarrow 70.13\%$). We postulate that the reason might be related to the forgetting issue \cite{chen2019catastrophic,boschini2022transfer,jung2016less} in the fine-tuning process. ProGrad is designed to avoid the forgetting issue by aligning the gradient descent direction of matching predictions and ground-truth labels to that of knowledge distillation when there is a conflict. ProGrad exhibits a good performance on that dataset.

We find that the performance difference between the DPL method and baseline methods is largest on the EuroSAT and Aircraft datasets. There is a large distribution shift on EuroSAT and Aircraft datasets compared to the pre-train datasets of the CLIP model. On the generic dataset such as Caltech101, the distribution shift is small. Hence, the difference between the DPL method and baseline methods is minimal.

\begin{figure}[t]
    \centering
    \includegraphics[width=\linewidth]{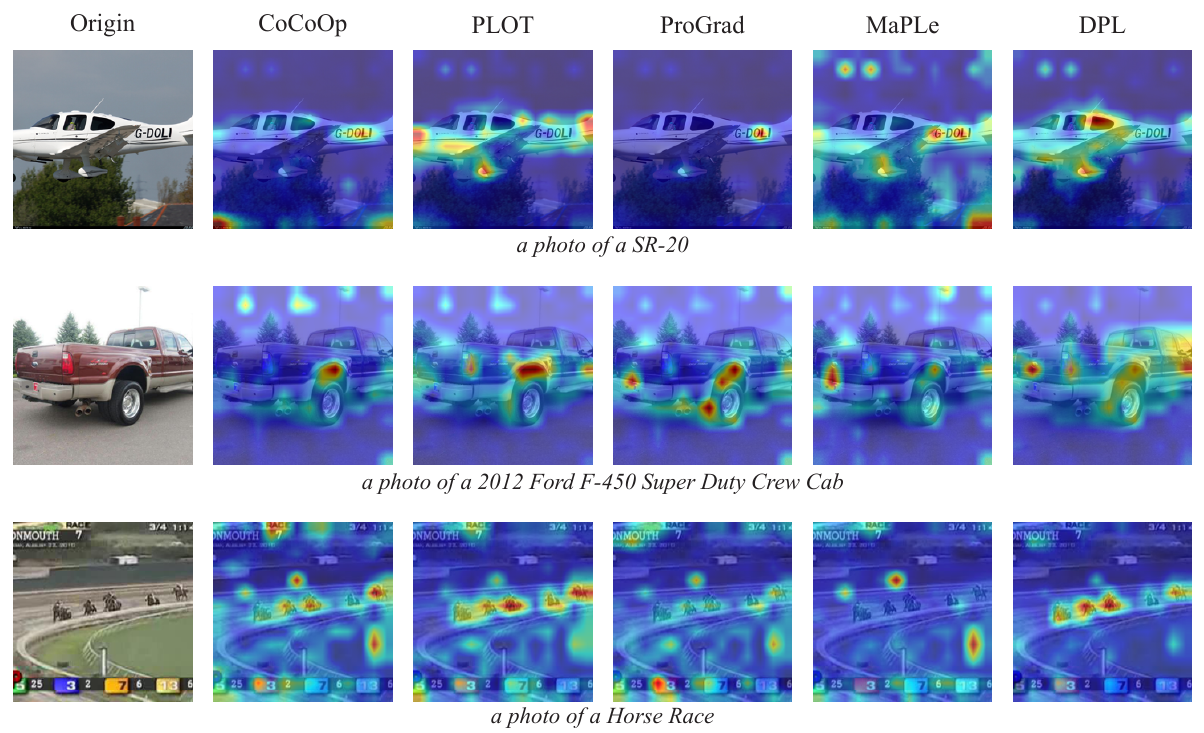}
    \caption[Caption]{Original image and Grad-CAM visualization \cite{selvaraju2017grad} for various methods on FGVCAircraft, StanfordCars and UCF101 datasets. Our method helps the pre-trained model focus on key elements in the foreground and avoid distraction from the background. The text template \texttt{a photo of a [class]} is used in the Grad-CAM calculation.}
    \label{fig:subnet_gradcam}
\end{figure}

We visualize the attention map on the downstream datasets using different prompt learning methods as shown in Figure \ref{fig:subnet_gradcam}. The Grad-CAM visualization \cite{selvaraju2017grad} indicates that our method is beneficial to focus on key elements in the foreground. In the image of amphibious aircraft, the attention of the model using the DPL method is localized on the float and the window. The model's attention focuses on the vehicle's logo in the car image. In the action classification, the model focuses on riders and horses. Those elements are important to differentiate the target class from the candidate classes. The PLOT method \cite{chen2022plot} is designed to align the prompts with image features using the optimal transport theory \cite{monge1781memoire,rubner2000earth,peyre2019computational}. It also exhibits a good alignment. The DPL method only relies on the prompt configuration, which means it is compatible with using the optimal transport theory.

\subsection{Ablation Studies}



We examine the performance of the shallow prompt learning method: continuous prompts with a context length of 16 are added to the inputs of the text branch and the image branch respectively. The equivalent prompt depth is 1. The way of adding context prompts is illustrated in Equation \ref{eq:cross_att_qkv} and Equation \ref{eq:cross_att_multi}. The result is shown in the Table \ref{table:prompt_ablation}. Compared to the DPL method, the shallow prompt learning method has a remarkable performance drop. There is the largest performance degradation in the EuroSAT and Aircraft datasets. These two datasets also exhibit the advantage of the DPL method over baseline methods shown in the Table \ref{table:subnet_perf}.

\begin{table}[htb]
    \caption{Ablation study on prompt configuration, i.e. the context length and the depth of continuous prompts. The shallow prompting uses a prompt depth of 1 and the context length of 16. Test accuracy variation is reported compared to the optimal prompt configuration found by the DPL method. A negative value indicates the performance drop when using the shallow prompting method.}
    \label{table:prompt_ablation}
    \resizebox{\linewidth}{!}{
    \begin{tabular}{cccccc}
    \hline
     Method & Caltech101 & DTD & EuroSAT & Aircraft \\
    \hline
     Shallow Prompting & -0.37 & -3.13 & -7.27 & -12.14 \\
    \hline
    Method & Food101 & Flowers & Pets & Cars \\
    \hline
     Shallow Prompting & -0.80 & -1.06 & -1.16 & -3.27 \\
    \hline
     Method & UCF & SUN397 & ImageNet & Average \\
    \hline
     Shallow Prompting & -3.03 & -3.40 & -1.73 & -3.40 \\
    \hline
    \end{tabular}}
\end{table}

\section{Discussion}
\label{sec:discuss}

By using bi-level optimization, the DPL method is able to automatically determine the context length of the continuous prompt for each layer. It seems to indicate that the optimal prompt configuration might be dataset-dependent. The deep neural networks are found to be brittle to even small distribution shits between the pre-training and fine-tuning datasets \cite{recht2019imagenet,hendrycks2019benchmarking,koh2021wilds}. A dataset-dependent training scheme provides the flexibility to the level of distribution shift. Our results show that the few-shot learning with DPL can boost the downstream accuracy.

Existing deep prompt learning works hardly consider a granular level of adding continuous prompts: each layer has the same context length and a hyperarameter of prompt depth is empirically determined. A prompt depth $\ell_p$ smaller than the model depth $\ell$ indicates that the last few layers of the pre-trained model might be close to a minima for the downstream dataset. In the transfer learning, not all layer parameters are responsible for the distribution shift. There are numerous researches on training different layers differently to mitigate the forgetting issue \cite{niu2022efficient,toneva2018empirical,davari2022probing} caused by the distribution shift: fine-tuning on the target domain using gradual unfreezing \cite{howard2018universal,mukherjee2019distilling,romero2020targeted}, using different learning rate for different layers \cite{ro2021autolr,shen2021partial}, and training part of layers \cite{lee2022surgical,vettoruzzo2024advances}.

If some layer parameters are close to optimal in downstream datasets, we believe there is no need to add continuous prompts to those layers. The deep prompt methods using $\ell_p < \ell$ avoid adding prompts to layers that are close to the optimal. If we refine this strategy, a natural question is \textit{do all $\ell$ layers require the same context length}? Different layers might have a different level of deviation from the minima. The extreme case is that a layer is very close to the minima and no continuous prompts are needed, i.e. the context length is 0 in this layer. For layers with different levels of deviation, we might need to use different context lengths. This is consistent with the result of using the DPL method. Overall, the result reveals two things: (1) the optimal context length depends on the distribution shift. (2) different layers of the pre-trained model might require different context lengths.

\section{Limitation}

The searching stage of the DPL method, similar to NAS, is computationally costly due to the introduction of the supprompt. The computational cost is determined by the size of the search space. At each depth, DPL searches context lengths of $\{0, 2, 4, 6\}$. The size of the search space is $2.81 \times 10^{15}$. Using differentiable approach greatly accelerates the searching process compared to exhaustive search. The training stage of the DPL method is lightweight as it relies completely on prompt configuration without advanced technologies such as coupling functions. Details regarding to the computational costs are discussed in Appendix \ref{append:complexity}.

\section{Conclusion}
\label{sec:conclusion}

In this work, we automate the prompt learning design by relaxing the categorical selection of context lengths to obtain a continuous search space. Using limited data, our method is able to find a prompt learning setting that is better than the existing manually designed prompt learning methods. The method is simple yet effective: it only focuses on 
determine context lengths of continuous prompts. We empirically find that our searching algorithm has a higher confidence in the text branch compared to that in the image branch, and that the prompt configuration shows a data-dependent behavior. The data-dependent behavior and the asymmetric prompt insertion in two branches demonstrate the strength of the automatic prompt learning design. In summary, our work proposes a new paradigm of adding continuous prompts that removes the restriction of the manually designed context length fixed prompt learning method.

\bibliographystyle{named}
\bibliography{ref}






\newpage
\appendix

\section{Appendix}
\subsection{Alpha Matrix Evolution}
\label{append:alpha_evolve}

\begin{figure*}[htb]
    \centering
    \includegraphics[width=\linewidth]{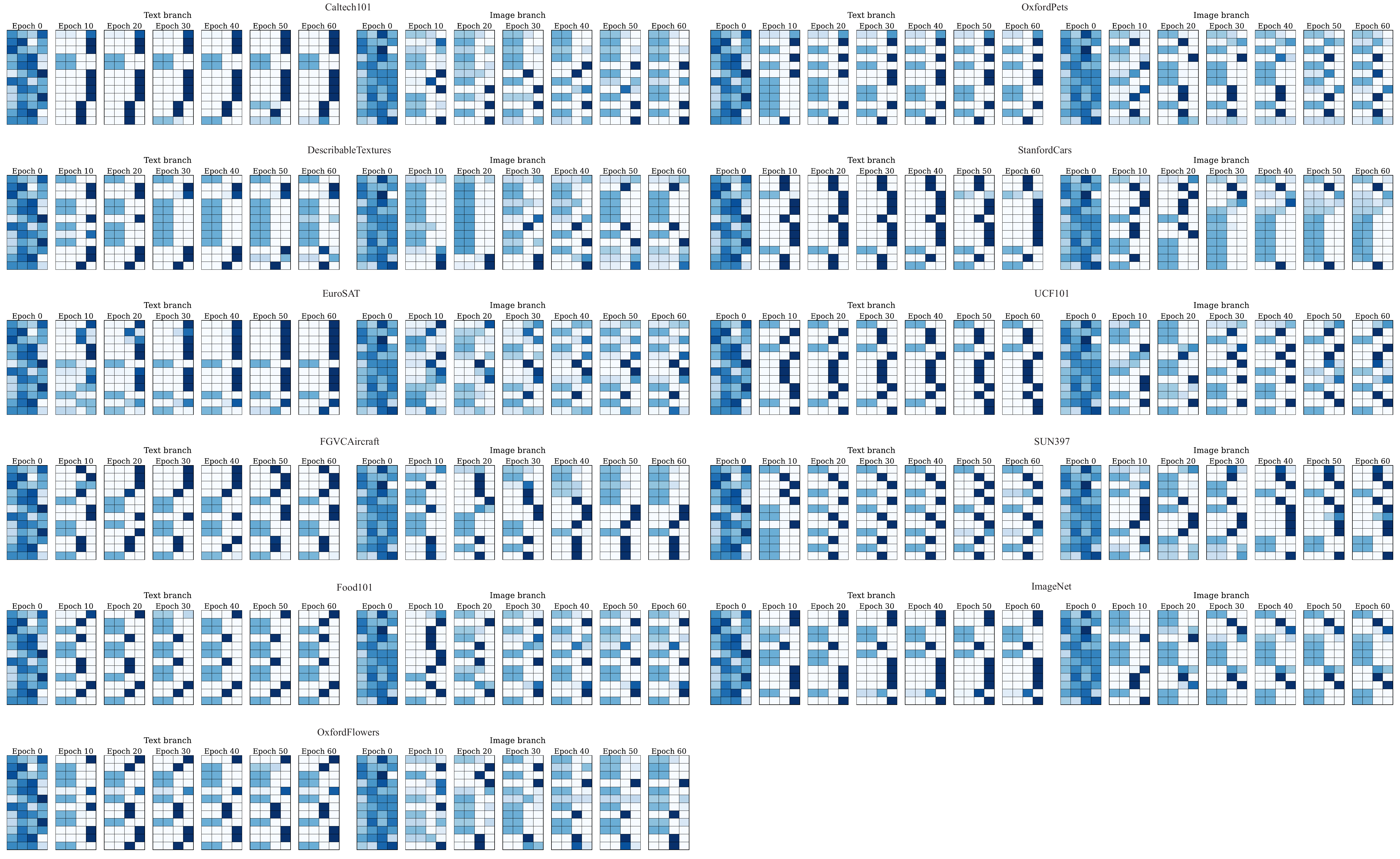}
    \caption{Evolution of $\alpha$ matrices using various datasets in the searching stage. Although $\alpha$ matrices have the same random initialization for the same text branch or image branch, the converged matrices are different for different datasets. It indicates that the prompt learning method depends on the distribution shift but the existing manually designed prompt learning method uses the same context length for different downstream dataets.}
    \label{fig:alpha_mat_all}
\end{figure*}

We examine the evolution of $\alpha$ matrix on the Caltech101, DescribableTextures, EuroSAT, FGVCAircraft, Food101, OxfordFlowers, OxfordPets, StanfordCars, UCF101, SUN397 and ImageNet datasets. The result is shown in Figure \ref{fig:alpha_mat_all}. Overall, the searching algorithm has a higher confidence in the text branch compared to the image branch. Even though using only 16 shots learning, the searching algorithm can find subprompts with reasonable confidence on all 11 datasets.

We note that the $\alpha$ matrix at the last epoch varies for the different datasets. It indicates the dataset-dependent behavior of the deep continuous prompt method.

\begin{table}[htb]
    \centering
    \caption{Terminologies and explanations used in this work.}
    \label{table:terminology}
    \begin{tabular}{
    >{\raggedright\arraybackslash}p{.26\linewidth}%
    >{\raggedright\arraybackslash}p{.64\linewidth}}
        \hline
        Terminologies & Explanations \\
        \hline
        \emph{Prompt configuration} & Prompt configuration specifies the context length of continuous prompts added to each layer. For example, in the shallow prompt learning, the prompt configuration is to add a continuous prompt with a length of $c_p$, $\mathbf{E} \in \mathbb{R}^{c_p \times d}$, only to the input. \\
        \emph{Search space} $\mathcal{A}$ & $\mathcal{A}$ contains all possible combination of adding continuous prompts. For an $\alpha$ matrix $\mathbf{A}^{\alpha} \in \mathbb{R}^{\ell \times t}$. The search space size is $\abs{\mathbf{A}^{\alpha}} = t^{\ell}$\\
        \emph{Supprompt} & A supprompt consists of a $\alpha$ matrix $\mathbf{A}^{\alpha} \in \mathbb{R}^{\ell \times t}$ and continuous prompts $\{ \mathbf{E}^{(l)}_{i} \in \mathbb{R}^{c_i \times d} \mid 1 \le i \le t, 1 \le l \le \ell, i, l \in \mathbb{N}^{+} \}$\\
        \emph{Subprompt} & A subprompt consists of continuous prompts $\{\mathbf{E}^{(l)} \in \mathbb{R}^{c_l \times d} \mid 1 \le l \le \ell, l \in \mathbb{N}^{+} \}$\\
        \emph{Confidence} & After the searching stage, an $\alpha$ matrix is said to have high confidence if the majority of the rows in the $\alpha$ matrix satisfies $\forall m, n, 1 \le m \le t, 1 \le n \le n, m \ne n, \abs{A^{\alpha}_{im} - A^{\alpha}_{in}} > T$, where $T$ is a threshold value. \\
        \hline
    \end{tabular}
\end{table}

\subsection{Terminology}
\label{append:terminology}

We summarize the terminology in Table \ref{table:terminology}. \textit{Supprompt} is analogous to Supernet while \textit{Subprompt} is analogous to Subprompt in differentiable NAS \cite{liu2018darts,xu2019pc,dong2019searching,liang2019darts+,zela2019understanding,chu2020fair,yan2021zeronas}. Similarly, we use \textit{search space} to group prompt configurations in the searching process. The computational cost of the searching stage is related to the size of the search space.

\textit{Prompt configuration} determines context lengths at different depths. Manually designed prompts generally have a homogeneous prompt configuration while the DPL method intentionally introduces heterogeneity in the prompt configuration. We use \textit{Confidence} to indicate the quality of the searching stage. When the confidence level is low, there are more than one options with high $\alpha$ values at various depths. Only one option with the highest $\alpha$ value is selected, so multiple options with similar $\alpha$ values can lead to the selection of the suboptimal prompt configuration. Hence, a high confidence value is preferred.

\subsection{Computational Complexity Comparison}
\label{append:complexity}

Table \ref{table:complexity} shows the computational complexity comparison. In the searching stage, the DPL method, similar to the differentiable NAS, has a larger number of trainable parameters as the supprompt incorporates all candidate prompt configurations.

In the training stage, the prompt configuration is data-dependent. Hence, the number of trainable parameters varies for different datasets. CoCoOp and MaPLe are using deep continuous prompts. Even though the supprompt in the searching stage has the largest number of trainable parameters, the computational complexity is smaller than the deep prompting method. Compared to the number of trainable parameters in the pretrained CLIP model, the number of parameters is much smaller in the prompt learning method. However, the downstream task performance is remarkably boosted compared to the zero-shot transfer, which demonstrates the strength of the PEFT methods.

\begin{figure}[htb]
    \centering
    \includegraphics[width=0.8\linewidth]{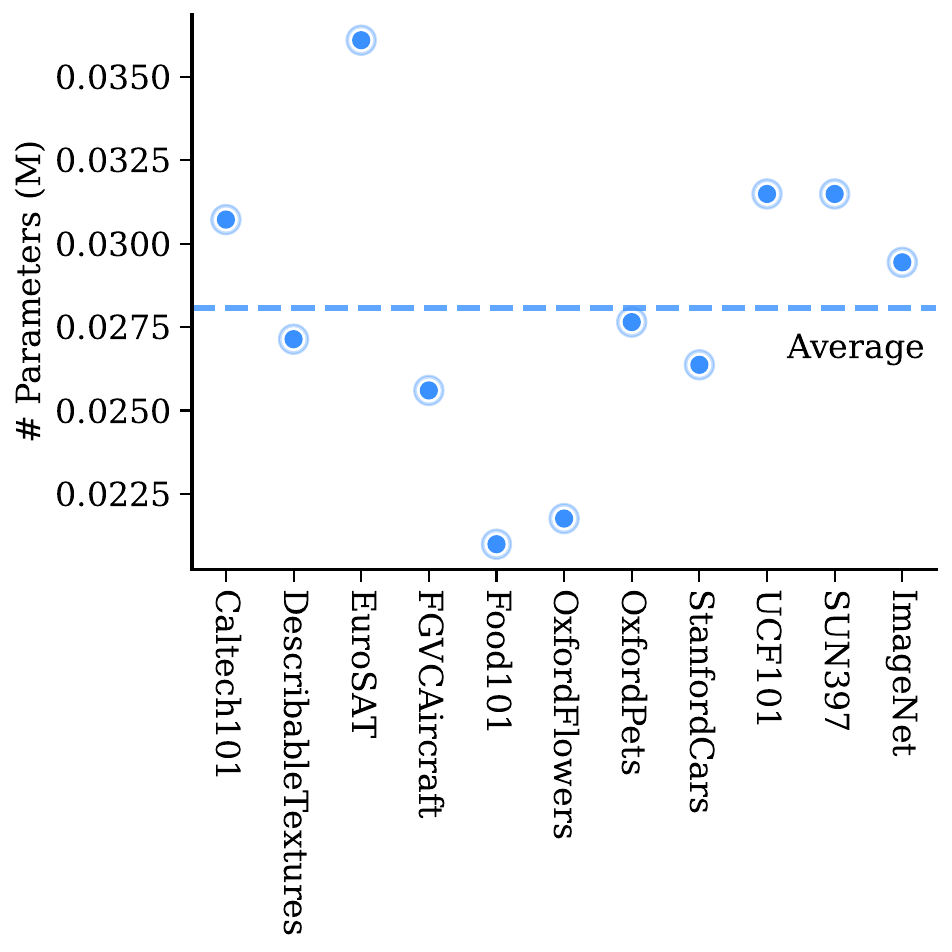}
    \caption{The computational complexity of the optimal prompt configuration determined by the DPL method.}
    \label{fig:param}
\end{figure}

The number of parameters in the training stage is in the range $(0, 0.12]$ M ($(0, 0.096]$\% of the original CLIP model parameter). The average number of parameters of the optimal prompt configurations determined by the DPL method is $0.028$ M. Figure \ref{fig:param} shows the number of trainable parameters on each dataset. Even the highest number of parameters in the training stage is smaller than the manually designed prompt method. At the same time, the DPL method can achieve the performance gain.
\begin{table}[htb]
    \caption{Computational complexity comparison between the DPL method and the baseline methods.}
    \label{table:complexity}
    \begin{center}
    \resizebox{.96\linewidth}{!}{
    \begin{tabular}{ccccc}
    \hline
    Type & Method  & Params & \makecell{Params \\\% CLIP} & Avg Test Acc \\
    \hline
    \multirow{3}{*}{Shallow prompt} & PLOT         & 8200        & 0.007      & 75.37 \\
     & ProGrad      & 8200        & 0.007      & 79.14 \\
     & CoCoOp       & 0.35 M      & 0.281      & 75.50 \\
    \hline
    \multirow{2}{*}{Deep prompt} & MaPLe        & 3.56 M      & 2.858      & 78.47 \\
    & DPL   & \makecell{0.23 M (Search) \\ 0.028 M (Train)}     & \makecell{0.185 (Search) \\ 0.022 (Train)}      & 81.71 \\
    \hline
    \end{tabular}}
    \end{center}
\end{table}

\begin{table}[htb]
    \caption{Test accuracies on downstream tasks with standard deviation reported using 16-shot learning. Experiments on every dataset are repeated over 3 runs.}
    \label{table:subnet_perf_std}
    \resizebox{\linewidth}{!}{
    \begin{tabular}{ccccccccc}
    \hline
    \rowcolor{gray!20} Method & Caltech101 & DTD & EuroSAT & Aircraft\\
    \hline
    ZS CLIP & 87.20 & 42.34 & 37.57 & 17.29 \\
    Linear Probe & 90.43 $\pm$ 0.21 & 64.03 $\pm$ 0.82 & 82.70 $\pm$ 1.06 & 36.37 $\pm$ 0.98 \\
    \hline
     CoCoOp  & 95.10 $\pm$ 0.08 & 63.63 $\pm$ 0.88 & 74.10 $\pm$ 0.57 & 33.67 $\pm$ 0.33 \\
     PLOT    & 93.70 $\pm$ 0.10 & 70.90 $\pm$ 0.54 & 84.03 $\pm$ 0.59 & 34.93 $\pm$ 1.05 \\
     ProGrad & 95.63 $\pm$ 0.39 & 66.27 $\pm$ 0.73 & 82.03 $\pm$ 1.52 & 41.30 $\pm$ 0.49 \\
     MaPLe   & 95.10 $\pm$ 0.16 & 67.27 $\pm$ 0.61 & 86.40 $\pm$ 1.47 & 37.07 $\pm$ 0.25 \\
     DPL    & 95.80 $\pm$ 0.24 & 71.50 $\pm$ 0.97 & 92.30 $\pm$ 0.18 & 48.57 $\pm$ 0.86 \\
     DPL + KD & 95.73 $\pm$ 0.19 & 70.90 $\pm$ 0.24 & 92.47 $\pm$ 0.77 & 49.43 $\pm$ 1.40 \\
     
     \hline
     \rowcolor{gray!20}Method & Food & Flowers & Pets & Cars \\
     \hline
     ZS CLIP & 77.30 & 66.18 & 85.79 & 55.63 \\
     Linear Probe & 70.13 $\pm$ 0.23 & 95.00 $\pm$ 0.67 & 76.37 $\pm$ 0.54 & 70.13 $\pm$ 0.61 \\
     \hline
     CoCoOp & 87.37 $\pm$ 0.12 & 89.97 $\pm$ 1.03 & 93.53 $\pm$ 0.45 & 72.30 $\pm$ 0.54 \\
     PLOT &  78.13 $\pm$ 0.21 & 97.27 $\pm$ 0.12 & 88.20 $\pm$ 0.51 & 68.10 $\pm$ 0.51 \\
     ProGrad & 86.70 $\pm$ 0.08 & 95.33 $\pm$ 0.38 & 93.10 $\pm$ 0.37 & 81.23 $\pm$ 0.57 \\
     MaPLe & 87.43 $\pm$ 0.09 & 94.27 $\pm$ 0.25 & 93.63 $\pm$ 0.34 & 74.87 $\pm$ 0.68 \\
     DPL & 82.57 $\pm$ 0.37	& 96.63 $\pm$ 0.53 & 92.03 $\pm$ 0.67 & 82.70 $\pm$ 0.72 \\
     DPL + KD & 82.80 $\pm$ 0.00 & 96.80 $\pm$ 0.24 & 91.83 $\pm$ 0.70 & 82.83 $\pm$ 0.11 \\
     
     \hline
     \rowcolor{gray!20}Method & UCF & SUN397 & ImageNet & \\
     \hline
     ZS CLIP & 61.45 & 58.73 & 58.77 \\
     Linear Probe & 73.70 $\pm$ 0.81 & 65.17 $\pm$ 0.32 & 54.33 $\pm$ 0.17 \\
     \hline
     CoCoOp & 76.97 $\pm$ 0.85 & 72.67 $\pm$ 0.05 & 71.17 $\pm$ 0.05 \\
     PLOT & 72.23 $\pm$ 0.12 & 69.40 $\pm$ 0.16 & 72.23 $\pm$ 0.12 \\
     ProGrad & 81.60 $\pm$ 0.71 & 75.13 $\pm$ 0.25 & 72.27 $\pm$ 0.05 \\
     MaPLe & 80.37 $\pm$ 0.78 & 74.73 $\pm$ 0.05 & 72.03 $\pm$ 0.12 \\
     DPL & 84.10 $\pm$ 0.65 & 79.83 $\pm$ 0.14 & 72.80 $\pm$ 0.32 \\
     DPL + KD & 83.77 $\pm$ 0.37 & 79.63 $\pm$ 0.30 & 73.03 $\pm$ 0.68 \\
     \hline
    \end{tabular}}
\end{table}

\subsection{Performance on Downstream Tasks}
\label{append:perf_downstream}
The standard deviation on each dataset is reported in Table \ref{table:subnet_perf_std}. Experiments are repeated over 3 runs and the number of shots is 16. Search space per depth is $\{0, 2, 4, 6\}$. We do not observe a large variation of the performance.

\subsection{Few-Shot Learning of the DPL method}
\label{append:few_shot}

\begin{figure*}[htb]
    \centering
    \includegraphics[width=.9\linewidth]{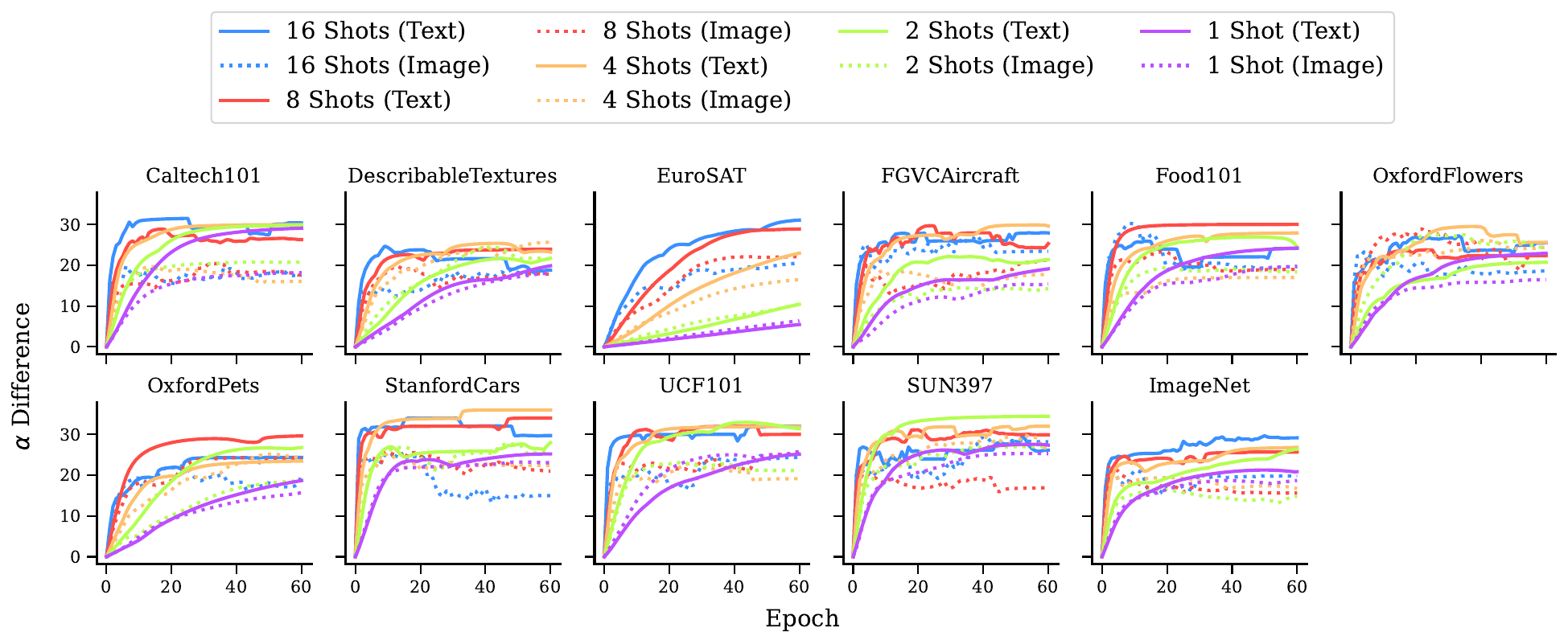}
    \caption{The variation of the $\alpha$ difference in the searching process. The number of shots is 16, 8, 4, 2 and 1. When using fewer shots in the searching stage, the searching algorithm becomes less confident. There is a pronounced drop in $\alpha$ difference using 1-shot learning.}
    \label{fig:few_shot_search}
\end{figure*}

We examine the performance of the DPL method using 16/8/4/2/1 shots. Figure \ref{fig:few_shot_search} shows the $\alpha$ difference on various datasets. The previous results have shown that the searching algorithm has higher confidence in the text branch compared to the image branch. When using fewer shots, the uncertainty increases. We note that on the EuroSAT dataset, the drop in the $\alpha$ difference is pronounced. The EuroSAT dataset contains satellite images that have a large distribution shift compared to the pre-trained dataset. Hence, it requires a larger number of shots.

When the number of shots is 1, there is a remarkable drop in $\alpha$ difference. The selection of context lengths is subjected to the perturbation and hence the searched prompt configuration might be away from the optimal.

\begin{figure*}[htb]
    \centering
    \includegraphics[width=\linewidth]{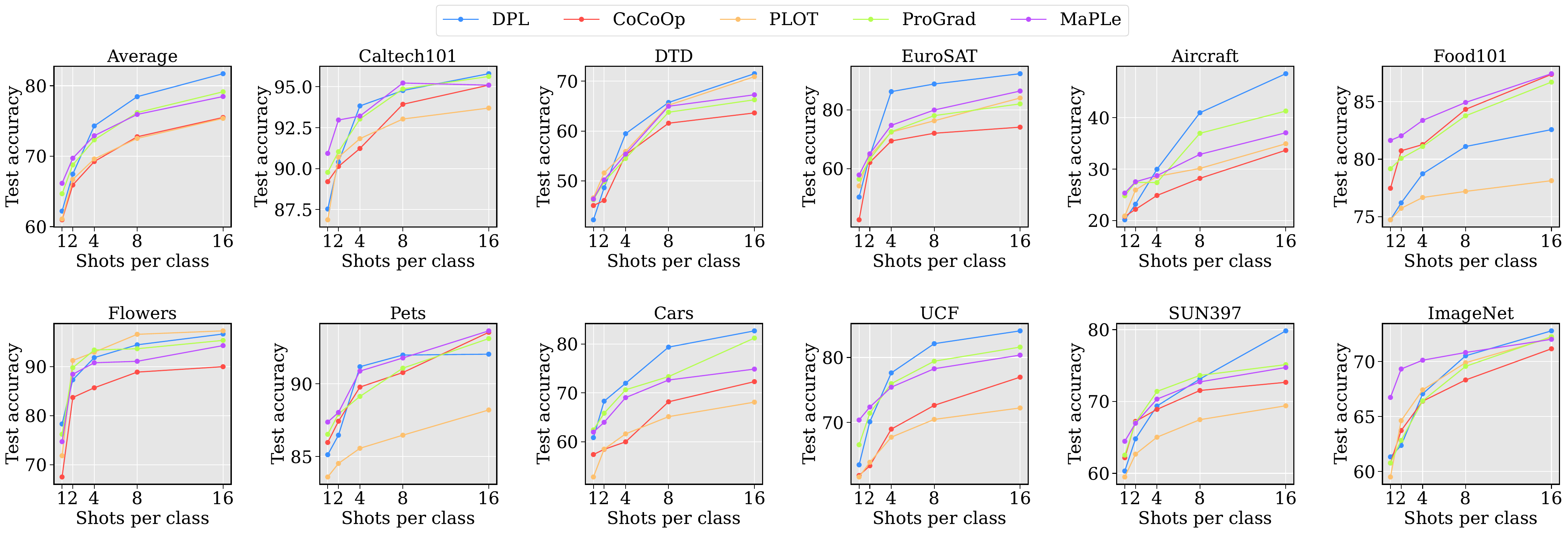}
    \caption{Performance of few-shot learning on 11 datasets for various prompting methods. The DPL method requires a fair amount of data (\eg 4, 8 and 16 shots) to exibit its advantage.}
    \label{fig:few_shot_train}
\end{figure*}

After determining the subprompt, we examine the performance in the few-shot learning setup (16/8/4/2/1 shots). The performance over 11 datasets is shown in Figure \ref{fig:few_shot_train}. When the number of shots is large, there is a significant performance boost for the DPL method. However, the DPL method becomes less competitive when the number of shots is very limited (\eg 1 and 2 shots). When the number of shots decreases, all prompting methods exhibit a conspicuous performance drop. Most prompting methods assume at least 16-shot data are available \cite{hirohashi2024prompt}. Similarly, the DPL method is not designed for very limited data. As indicated in the searching process in Figure \ref{fig:few_shot_search}, limited data can cause the searched subprompt to be suboptimal. Hence, training the suboptimal subprompt can lead to unfavorable performance. In other words, the bi-level optimization process requires a fair amount of data for better convergence.

\end{document}